\documentclass[10pt,twocolumn,letterpaper]{article}

\newif\ifarxiv
\newif\ifcvpr
\newif\ifcvprfinal

\cvprfalse \cvprfinalfalse \arxivtrue %

\ifcvprfinal
\usepackage{cvpr} %
\fi
\ifcvpr
\usepackage[review]{cvpr} %
\fi
\ifarxiv
\usepackage[pagenumbers]{cvpr} %
\fi

\usepackage[pagebackref,breaklinks,colorlinks,allcolors=blue]{hyperref}
\usepackage[table]{xcolor}

\def\paperID{18495} 
\def\confName{CVPR}
\def\confYear{2025}
\def\CoolName{SplatR }

\title{\CoolName : Experience Goal Visual Rearrangement with 3D Gaussian Splatting and Dense Feature Matching}

\author{Arjun P S\\
IIT ISM Dhanbad\\
{\tt\small papers.arjun@gmail.com}
\and
Andrew Melnik\\
University of Bremen\\
\and
Gora Chand Nandi\\
IIIT Allahabad\\
}

\begin{document}

\twocolumn[{
\maketitle
\vspace{-1.2em}
\begin{center}
    \centering
    \includegraphics[width=1\linewidth]{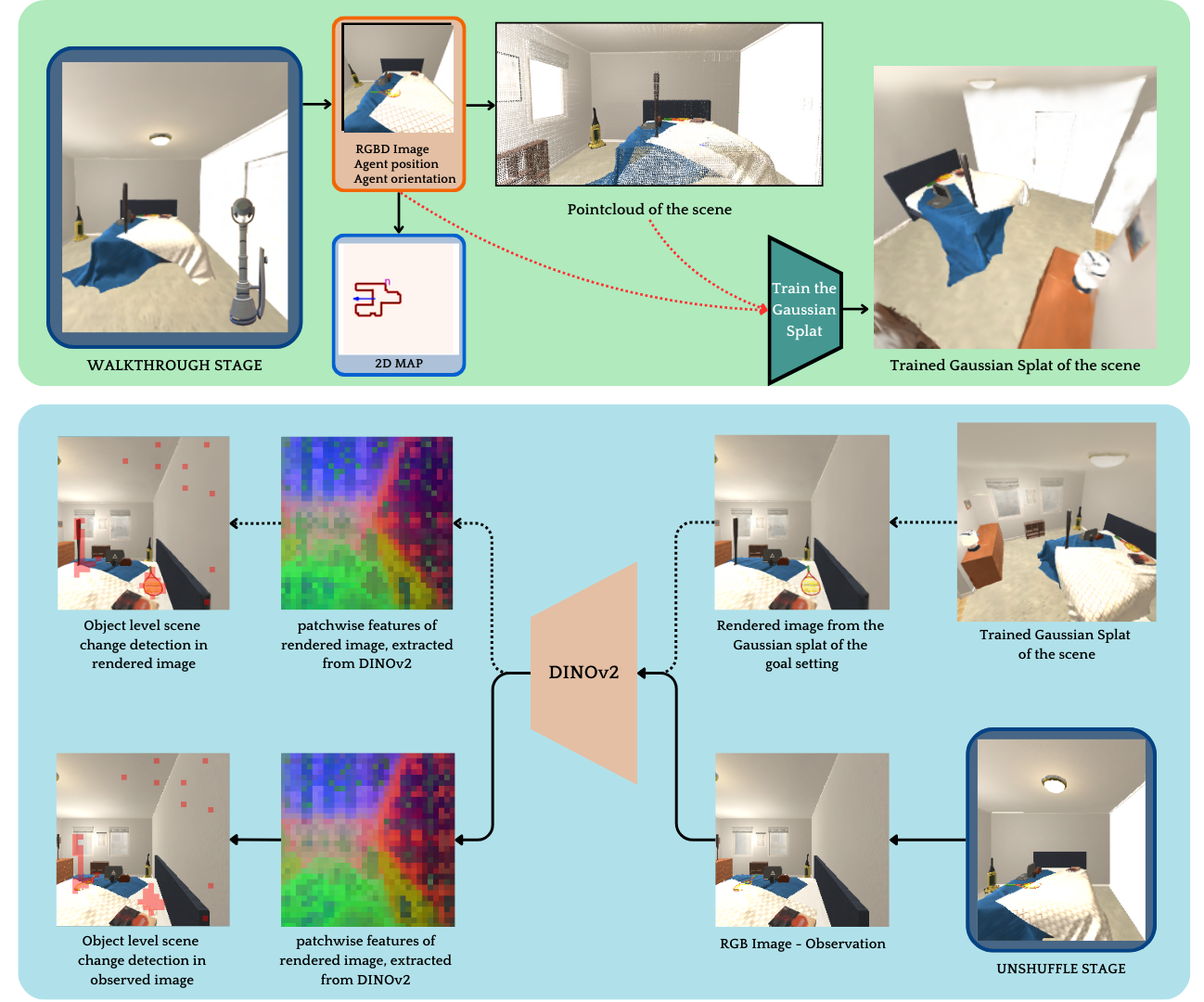}
    \captionof{figure}{\textbf{\CoolName}is an Embodied AI agent, that solves the experience goal rearrangement task by building a 3D Gaussian splat as a 3D scene representation. The agent initialized in the goal setting, collects observation and builds the Gaussian Splat to save the goal configuration. Reintroduced into the same world with shuffled object configuration, \CoolName explores the scene and renders a consistent view from the Gaussian Splat. Changes in the scene are detected by the similarity between corresponding patchwise features extracted from DINOv2.}
    \label{fig:pipeline}
\end{center}
}]

\begin{abstract}
Experience Goal Visual Rearrangement task  stands as a foundational challenge within Embodied AI, requiring an agent to construct a robust world model that accurately captures the goal state. The agent uses this world model to restore a shuffled scene to its original configuration, making an accurate representation of the world essential for successfully completing the task. In this work, we present a novel framework that leverages on 3D Gaussian Splatting as a 3D scene representation for experience goal visual rearrangement task. Recent advances in volumetric scene representation like 3D Gaussian Splatting, offer fast rendering of high quality and photo-realistic novel views. Our approach enables the agent to have consistent views of the current and the goal setting of the rearrangement task, which enables the agent to directly compare the goal state and the shuffled state of the world in image space. To compare these views, we propose to use a dense feature matching method with visual features extracted from a foundation model, leveraging its advantages of a more universal feature representation, which facilitates robustness, and generalization. We validate our approach on the AI2-THOR rearrangement challenge benchmark and demonstrate improvements over the current state-of-the-art methods.
\newline
Project Webpage: \href{https://splat-r.github.io/}{https://splat-r.github.io/}
\end{abstract}
\section{Introduction}

Rearrangement challenge \cite{batra2020rearrangementchallengeembodiedai} is a pivotal benchmark in developing Embodied AI agents that can interact with the physical world. It involves navigating through complex scenes, exploring and recognizing the current state of the world, reasoning about the changes in the world from a goal state and manipulating objects in the regions of change to bring it to the goal state where the goal specification can be represented in terms of a geometric transformation, image, language, experience or a predicate. This work focuses on the experience goal setting, where the agent is immersed in the environment at its goal state, letting the agent build a representation of the world. The agent is then initialized in a shuffled state of the same environment and is tasked to rearrange it to the goal state.

The most significant parts of the rearrangement task with an experience goal setting are the effectiveness of the representation built when the agent is exposed to the goal setting and the capability of the agent to compare the new state configuration with the retrieved goal state configuration. The model of the environment should be capable of encoding the intricate semantics within the scene. Prior works have focused on modelling the scene with a variety of methods \cite{yenamandra2024towards, melnik2023uniteam}: scene graphs where nodes are objects and the relation between objects are represented with edges \cite{sarch2022tideetidyingnovelrooms, gadre2022continuousscenerepresentationsembodied}, 3D voxelized maps \cite{trabucco2022simpleapproachvisualrearrangement}, 2D semantic maps \cite{sarch2022tideetidyingnovelrooms, sheikh2023language}, pointcloud based representation \cite{Liu_2024_CVPR} and images \cite{weihs2021visualroomrearrangement}. 

However, these methods have disadvantages: 2D and 3D semantic maps store object pose and semantic information in a grid; this approach provides limited resolution, does not inherently capture interactions between objects and is prone to sensitivity issues and quantization errors. Although pointcloud based representation can provide more robustness to sensitivity, it lacks structural semantics: identifying objects and their interactions with the world in a noisy pointcloud is challenging. Scene graph based methods often assume a clear and well defined relationship between objects, which often limits the granularity of scene understanding, and the abstract, ambiguous and complex interrelations between all the objects in the scene can be difficult to capture in a rigid graph structure. Images, on the other hand, provides rich visual information about the scene but lack continuity across the scene. Since images are taken from discrete locations, they fail to provide continuous information about the scene and, storing only limited viewpoints makes it harder to infer the intricate dynamics of a scene. To compare changes in the scene, the agent would need to return to the exact location and viewpoint to capture a consistent image, which may not be possible in a complex and cluttered scene. 

In this context, our work aimed to use 3D Gaussian Splatting \cite{kerbl3Dgaussians, guédon2023sugarsurfacealignedgaussiansplatting} as a scene representation for the rearrangement task. 3D Gaussian splatting based volumetric representation enables fast and real-time rendering of the scene by optimizing a set of 3D Gaussian primitives and is capable of synthesizing high quality and photorealistic novel views. By training a Gaussian Splat of the goal setting, the agent can freely explore the scene and render the corresponding viewpoint from the splat using a virtual camera, thus enabling the agent to have consistent view of the current and goal setting. This volumetric representation provides us with the benefit of both images and pointclouds: continuous scene representation with high quality rendered images that provides rich visual information regarding the scene. 

Consistent images from the current observation frame and rendered view enable a direct comparison. Even though the images are similar in their visual aspects, they are different at the pixel level. Image from the trained Gaussian Splat is rendered by rasterizing 3D Gaussians to a 2D plane, which might create artifacts in the rendered image. Considering this, We implement a patchwise dense feature matching method with patch-level features extracted from a foundational model like DINOv2 \cite{oquab2024dinov2learningrobustvisual} to compare the two images. Patchwise matching with visual features is more robust at handling variations than pixel-level comparison methods. Visual features from a foundation model like DINOv2 can capture semantically meaningful representations of the image patch. Changes in the scene can be recognized by matching the features of corresponding patches across the current and the rendered frame. patchwise changes in the images from the current and the goal setting are then grouped together to form object level masks. A category agnostic matching with visual embedding is performed to match objects in the current configuration with those in the goal configuration. 

We validate our method to perform the experience goal based rearrangement task using the AI2THOR Rearrangement Challenge \cite{weihs2021visualroomrearrangement} that utilizes the Room Rearrangement (RoomR) dataset. The challenge has two phases: In the Walkthrough phase, the agent is initialized in a room at a certain state, and it explores and builds a representation of the scene; In the Unshuffle phase, the agent is initialized in the same room with shuffled object states and the agent is tasked to restore the object states to the goal configuration observed during the walkthrough phase. Our approach shows improvement over the current state of the art methods for experience goal rearrangement task.
\section{Related Works}
\label{sec:related_works}

\textbf{Visual Rearrangement}; General rearrangement problem involves manipulating objects in the scene, to change its current state to a goal state. Our work focuses on a subset of this problem that involves exposing the agent to the goal setting before shuffling the environment and asking the agent to bring it back to its original state. Prior works used end-to-end methods \cite{weihs2021visualroomrearrangement}, scene graph based planning \cite{gadre2022continuousscenerepresentationsembodied}, 3D voxel grid based methods \cite{trabucco2022simpleapproachvisualrearrangement}, combination of 2D semantic map and memory graph \cite{sarch2022tideetidyingnovelrooms} and pointcloud based matching \cite{Liu_2024_CVPR} to perform experience goal rearrangement task. In detail, \cite{weihs2021visualroomrearrangement} employs a combination of CNN's and embedding layers to process input image, and an RNN module to reason about the processed information. \cite{gadre2022continuousscenerepresentationsembodied} encodes interaction between objects into a graph, where the edges are encoded with an embedding vector. This continuous scene representation is used for detecting changes in the scene. \cite{sarch2022tideetidyingnovelrooms} focuses on using a modular approach to solve the rearrangement task. It uses an out-of-place detection module that uses visual features and language descriptions of the object and its interrelations with the scene to detect whether it is out of place or not, memory graph network to store goal configuration and a visual search network to explore the scene. In contrast to the above mentioned works that uses an object detection module, \cite{Liu_2024_CVPR} uses a category agnostic matching of pointcloud representations created by comparing differences in pointcloud in the shuffled and goal configuration.

\textbf{3D Gaussian splatting} \cite{kerbl3Dgaussians} has emerged as a prominent method in novel view synthesis and 3D scene reconstruction due to is fast differentiable rendering capabilities by optimizing a set of 3D Gaussian primitives and splatting them to a 2D plane. Works like \cite{xie2024physgaussianphysicsintegrated3dgaussians, feng2024gaussiansplashingunifiedparticles, waczyńska2024gamesmeshbasedadaptingmodification, zhong2024reconstructionsimulationelasticobjects, qiu2024featuresplattinglanguagedrivenphysicsbased} used 3D Gaussian splatting for modelling underlying physics and dynamics of the scene, \cite{wu20244dgaussiansplattingrealtime, yang2024realtimephotorealisticdynamicscene,duan20244drotorgaussiansplattingefficient, gao2024gaussianflowsplattinggaussiandynamics, xiao2024bridging3dgaussianmesh, chu2024dreamscene4ddynamicmultiobjectscene, lu20243dgeometryawaredeformablegaussian, li2024spacetimegaussianfeaturesplatting} extended 3D Gaussian Splatting to dynamic scenes, \cite{guédon2023sugarsurfacealignedgaussiansplatting, Huang_2024, wolf2024gs2meshsurfacereconstructiongaussian, fan2024trim3dgaussiansplatting} focused on geometrically accurate surface reconstruction from 3D Gaussian splatting and \cite{qin2024langsplat3dlanguagegaussian, shi2023languageembedded3dgaussians, zhou2024feature3dgssupercharging3d, zuo2024fmgsfoundationmodelembedded} emphasizes on embedding language and semantics to 3D Gaussians for 3D scene understanding. Subsequent works have also shown the utility of using 3D Gaussian Splatting for navigation \cite{lei2024gaussnavgaussiansplattingvisual, meng2024beingsbayesianembodiedimagegoal}, SLAM \cite{keetha2024splatamsplattrack, yan2024gsslamdensevisualslam, li2024sgsslamsemanticgaussiansplatting, zhu2024semgaussslamdensesemanticgaussian, huang2024photoslamrealtimesimultaneouslocalization, matsuki2024gaussiansplattingslam} and manipulation \cite{abouchakra2024physicallyembodiedgaussiansplatting, lu2024manigaussiandynamicgaussiansplatting, shorinwa2024splatmovermultistageopenvocabularyrobotic}.
\section{Method}
\label{sec:method}

\subsection{Experience Goal Visual Rearrangement}
Visual Rearrangement in general follows Partially Observable Markov Decision Processes (POMDPs) \cite{batra2020rearrangementchallengeembodiedai}, where the agent no access to the state space $S$. State space of the world $S$ in rearrangement is generally represented as the Cartesian product of states of all the rigid bodies $S_i$ within the world. Each state of a rigid body can be represented as $S_i = \mathbb{R}^3 \times SO(3)$, where $\mathbb{R}^3$ represents the position and special orthogonal group $SO(3)$ represents the orientation of the rigid body. The state space for a world with $n$ objects,
\begin{equation}
    S = (\mathbb{R}^3 \times SO(3))_1 \times (\mathbb{R}^3 \times SO(3))_2 \dots (\mathbb{R}^3 \times SO(3))_n
\end{equation}

\subsection{Overview}
\CoolName is the first method to use Gaussian Splatting for experience goal rearrangement task. During the first phase of the task, the agent is immersed in a scene where the objects are in the goal configuration. The agent explores the environment to collect data from multiple viewpoints (\cref{subsec:data_collection}). The agent uses this data to train a Gaussian Splat of the goal setting (\cref{subsec:gaussian_splat}). In the second phase of the task, the agent is initialized in the same scene with shuffled object states. As the agent explores the shuffled scene, the agent uses a virtual camera in the trained Gaussian Splat to render images of the goal configuration. By using patch-wise dense feature matching with DINOv2 (\cref{subsec:feat_matching}), the agent recognizes the regions in the current observed image, which differ from the goal setting and vice-versa. These regions and their spatial and semantic information are stored separately as objects in the current and goal setting. At the end of exploration, the agent uses a category agnostic semantic matching method (\cref{subsec:cat_agn_matching}), to find objects in the current scene configuration that are similar to objects in the goal configuration. The agent uses these matched object pairs to perform the rearrangement task. Overall pipeline of \CoolName is presented in \cref{fig:pipeline}.

\subsection{Exploration and Data Collection}
\label{subsec:data_collection}

Intelligent exploration is crucial in collecting high quality data for training the Gaussian Splat. The agent must ensure that the scene is fully explored and well observed to avoid any artifacts. To ensure this, \CoolName incrementally builds a 2D map of the scene, we leverage the 2D map module from \cite{sarch2022tideetidyingnovelrooms} for keeping track of obstacles in the scene. The exploration strategy is designed in such a way that the agent covers the entire scene. This is done by randomly sampling a point on the unexplored traversable regions of the map at that timestep, using a fast-marching method \cite{doi:10.1073/pnas.93.4.1591} to navigate to this point and allowing the agent to add increments to the map, as it explores new regions of the scene. This is continued until all the traversible regions have been explored. The obstacle map is built only when the agent finds an obstacle in front of it. This ensures that all the obstacles in the obstacle map are built when the agent is close to those obstacles, thus ensuring close-up views of objects.

For every timestep, the agent saves RGB image, position and orientation of the agent with respect to the world frame, and the pointcloud data. At the end of exploration, the pointcloud is downsampled to a lower resolution by averaging points within a specified voxel grid size. This is done to limit the amount of Gaussians initialized during training, which reduces the overall computational overhead. 

\subsection{Gaussian Splatting}
\label{subsec:gaussian_splat}
\CoolName builds a Gaussian Splat based volumetric representation \cite{kerbl3Dgaussians} of the scene, to perform the rearrangement task. This representation is constructed by initializing a set of Gaussians in the scene and optimizing their parameters. Each Gaussian $g$ is parameterized by its mean $\mu_g \in \mathbb{R}^3$, 3D covariance $\Sigma_g$ that is parameterized by a scaling vector $s_g \in \mathbb{R}^3$ and rotation matrix $R_g$, opacity $\alpha_g \in [0, 1]$ and RGB color $C$ which is represented by a set of spherical harmonic coefficients. For a point $x \in \mathbb{R}^3$ in 3D space, the 3D Gaussian equation (\cref{eq:gauss_fn}) and the parameterized covariance (\cref{eq:cov}) :
\begin{equation}
	G_g(x) = e^{-\frac{1}{2}(x - \mu_g)^{T}\Sigma_g^{-1}(x - \mu_g)}
    \label{eq:gauss_fn}
\end{equation}

\begin{equation}
	\Sigma = RSS^TR^T
    \label{eq:cov}
\end{equation}

Gaussian splatting uses differentiable rendering with a tile-based rasterizer \cite{kerbl3Dgaussians} to rasterize 3D Gaussians to a 2D image plane \cite{964490}. The rendered color of a pixel $p = (u, v)$ in the 2D image (\cref{eq:render}) :
\begin{equation}
    C(p) = \sum_{i \in \mathcal{N}} c_i \alpha_i G_i^{2D}(p) \prod_{j=1}^{i-1} (1 - \alpha_j G_j^{2D}(p))
    \label{eq:render}
\end{equation}

where $\mathcal{N}$ is the number of Gaussians in the tile, $c_i$ is the color of the $i$-{th} Gaussian and $G_i^{2D}(p)$ is the 2D projection of 3D Gaussian $G_i(x)$ in equation \cref{eq:gauss_fn}, where $G^{2D}(.)$ is parameterized by the covariance matrix in camera coordinate frame $\Sigma'$. According to \cite{kerbl3Dgaussians}, $\Sigma'$ is computed using the viewing transformation matrix $W$ and the Jacobian $J$, which is the affine approximation of the projective transformation.
\begin{equation}
	\Sigma' = J W ~\Sigma ~W ^{T}J^{T}
     \label{eq:raster_gauss}
\end{equation}

\begin{figure*}
    \centering
    \includegraphics[width=1\linewidth]{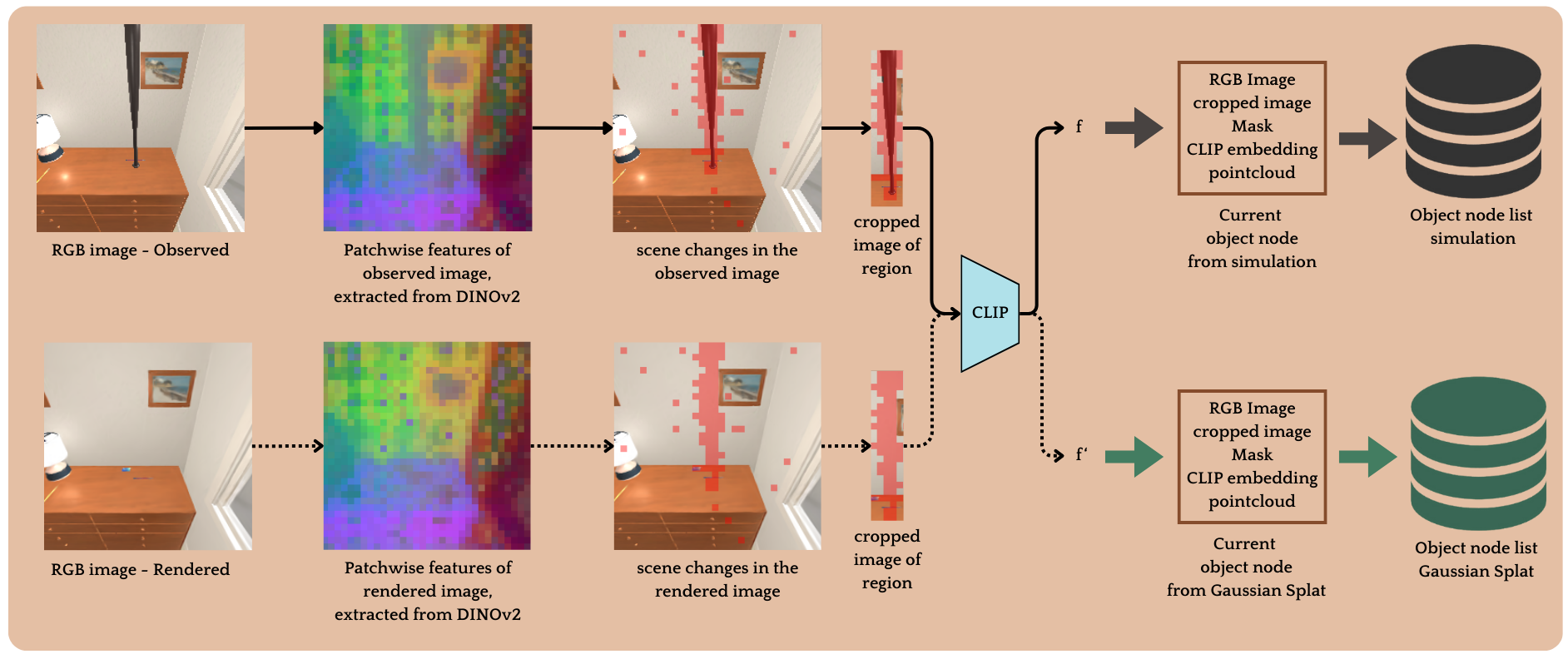}
    \caption{Overview of the scene change detection and storage framework. Images, observed by the agent and rendered from the Gaussian Splat are compared with a patchwise feature matching method. The resulting detections are stored as an object node. The patchwise feature visualization above is generated by taking the PCA (principal component analysis) of combined features in image of the current and goal setting.}
    \label{fig:scm_pipeline}
\end{figure*}

After training the Gaussian Splat, the agent is initialized in the same scene with shuffled object configuration. We use the camera intrinsics and the current pose of the agent to initialize a virtual camera in the Gaussian Splat that views the world from the same viewpoint as that of the agent. The virtual camera is designed to follow the same trajectory as that of the agent in the shuffled scene, so at any timestep, the agent has the currently observed image and the corresponding consistent image of the scene before shuffling the object states. This allows the agent to make a direct comparison between the images. An overview of the scene change detection pipeline is presented in \cref{fig:scm_pipeline}

\subsection{Patch-Wise Dense Feature Matching}
\label{subsec:feat_matching}
We utilize patch-level features extracted from the DINOv2 \cite{oquab2024dinov2learningrobustvisual} model to recognize changes in the image. We selected a foundational model like DINOv2 over the latent feature representation of a task specific model trained on the dataset used in this work, to enhance transferability across diverse datasets and real-world applications. DINOv2's task-agnostic training on a large and diverse corpus endows it with robust generalization capabilities. DINOv2 follows a vision transformer architecture where an image input returns a class token, patch tokens with patch size 14 and optionally 4 register tokens. Images from the current $I^C$ and goal $I^G$ setting are passed through the DINOv2 model to get their corresponding patch-wise features $f^C$ and $f^G$. We compute the cosine similarity $S_{ij}$ between extracted features $f_{ij}^C$ and $f_{ij}^G$ of corresponding patch $(i, j)$ in the current and goal configuration, to determine if those patches are similar or not. We group adjacent similar patches, that are dissimilar across the current and rendered image to form objects.
\begin{equation}
    S_{ij} = \frac{f_{ij}^C \cdot f_{ij}^G}{\|f_{ij}^C\| \|f_{ij}^G\|}
    \label{eq:sim_patch}
\end{equation}

The detections retrieved from the patchwise feature matching method can often be noisy. This is generally due to artifacts in the trained Gaussian Splat. It has been observed that these artifacts generally occur on plain or reflecting surfaces like walls or mirrors. Since the agent has access to all the object names in the RoomR dataset, to address the noisy detections, the agent computes the similarity (cosine similarity of CLIP feature vectors of the region and object names) of the detected region with the list of objects conditioned with the words "wall" and "mirror". If the computed similarity is highest for the conditioned terms, it will likely be a false detection.

\subsection{Object Nodes}
\label{subsec:obj_nodes}

For every object level detection, \CoolName stores an object node for the shuffled and goal setting separately. This object node consist of the corresponding image $I$, mask $m$, semantic information: CLIP \cite{radford2021learningtransferablevisualmodels} embedding $g$ of an image cropped along the objects mask and spatial information: center coordinate $p$ and pointcloud $P$ in world space, of the object. Since we use patchwise features, the masks for objects in the scene are not accurate. To address this, we use segment anything model \cite{kirillov2023segany} (SAM) to find the accurate mask, by prompting the image with a bounding box prompt, where the bounding box is along the mask obtained from grouped DINOv2 patches.

Although an accurate mask is essential for manipulating misplaced objects, running SAM along with DINOv2 for every step is computationally inefficient. To optimize this, we store only masks derived from DINOv2 for each node and compute accurate masks for object nodes after the exploration phase.

\subsection{Merging Objects}
\label{subsec:obj_det}
 For each incoming object node, the agent calculates a similarity $\Phi^{node}$ index based on its visual $\Phi^{vis}$ and spatial similarity $\Phi^{sp}$ to the nodes stored in memory. For an incoming object node $O$ and a node $O_i$ in the memory (of $N$ nodes)
\begin{equation}
    \Phi_i^{node}(O, O_i) = \delta \Phi_i^{vis}(O, O_i) + (1-\delta)\Phi_i^{sp}(O, O_i)
    \label{eq:full_sim_node}
\end{equation}
where $\delta$ is a weighting factor.
\begin{equation}
    \Phi_i^{vis}(O, O_i) = \frac{g \cdot g_i}{\|g\| \|g_i\|}
    \label{eq:vis_sim_node}
\end{equation}
\begin{equation}
    \Phi_i^{sp}(O, O_i) = nnratio(P, P_i)
    \label{eq:sp_sim_node}
\end{equation}
where $nnratio$ \cite{gu2023conceptgraphsopenvocabulary3dscene} is the proportion of points in the pointcloud $P$, that have nearest neighbors in pointcloud $P_i$ of the $i$-th object $O_i$.

The similarity index $\Phi^{node}$ is used to match the incoming node with the existing nodes in memory. If the highest similarity between the incoming node and an existing node $O_j$ in memory exceeds a predefined threshold for $\tau_{sim}$, the new node is merged with the corresponding existing node $j$.
After merging, we store the information from both views of the same object. The new feature vector $g_j$ of the node $O_j$ after fusing it with incoming node $O$ of feature vector $g$ is given by,
\begin{equation}
    g_j = \frac{n_o^j \cdot g_j + g}{n_o^j + 1}
\end{equation}
Where $n_o^j$ is the number of merges that node $O_j$ has gone through prior to its merging with the current incoming node $O$.

\begin{figure}
    \centering
    \includegraphics[width=1\linewidth]{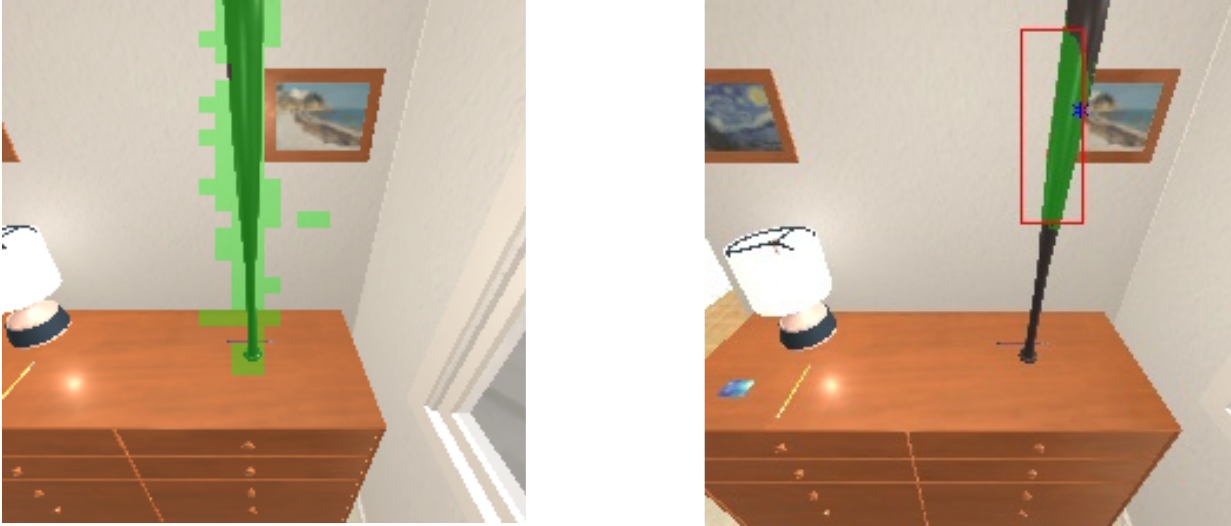}
    \caption{\textbf{Left}: Mask for an object generated by accumulating similar patches, that are dissimilar across the current and the goal setting. \textbf{Right}: Accurate mask obtained from SAM, for the same object observed during rearrangement.}
    \label{fig:sam_masks}
\end{figure}

\subsection{Category Agnostic Matching}
\label{subsec:cat_agn_matching}

The agent continues to build up the object level nodes until the end of exploration. After completely exploring the scene with shuffled object states, the agent uses the clip embedding $g$ stored in the object nodes $O$ for matching those in the current setting with the goal setting. Instead of classifying objects and matching them based on their classes, we directly compare the visual features of the objects. By doing so, we circumvent the potential inaccuracies associated with object detection.

At the end of exploration, we have a set of object nodes $\eta_s = \{O_s^1, O_s^2, \dots, O_s^1\}$ and $\eta_g = \{O_g^1, O_g^2, \dots, O_g^1\}$ corresponding to the shuffled configuration and goal configuration respectively. The matching between $\eta_s$ and $\eta_g$ can be considered as a weighted bipartite matching problem, where the weight is the cosine similarity between clip embeddings in the object nodes. For an object node $i$ in $\eta_s$ and an object node $j$ in $\eta_g$, the strength of edge between these two nodes,
\begin{equation}
    S^{obj}(O_i, O_j) = \frac{g_i^s \cdot g_j^g}{\|g_i^s\| \|g_j^g\|}
    \label{eq:sim_obj}
\end{equation}

We use the Hungarian algorithm (Kuhn Munkres algorithm) \cite{https://doi.org/10.1002/nav.3800020109}, to find the optimal matching in this bipartite graph. We have also experimented with a greedy assignment strategy, where each object node from the current setting is matched with the object node in the goal setting with the highest similarity (cosine similarity between feature vectors $f$ of both object nodes), in a sequential first come first serve manner.

\subsection{Rearranging the Scene}
\label{subsec:implementation_details}
Category agnostic matching provides the agent with a set of object node pairs, where each pair consists of a shuffled object node from the simulated scene matched to its corresponding configuration in the goal state. These matched pairs are ordered by similarity and processed sequentially to complete the rearrangement task.

Accurately picking up an object requires determining its precise position in 3D space. However, the patchwise masks generated by our pipeline are insufficient to capture the full geometric extent of an object. To obtain accurate masks, the image stored in the object node and with a bounding box prompt along the object's patchwise mask, is provided to SAM, which then returns a refined mask of the object. This refined mask is used for picking up the object. This process in visualized in \cref{fig:sam_masks}.
\section{Experimental Setup}
\label{sec:Experimental Setup}

\subsection{Rearrangement Task and Dataset}

We use the AI2-THOR rearrangement challenge as a benchmark \cite{weihs2021visualroomrearrangement} for evaluating our method. This challenge is based on the Room Rearrangement dataset (RoomR), built on AI2-THOR \cite{kolve2022ai2thorinteractive3denvironment} virtual environment. The RoomR dataset consist of rearrangement tasks with 120 rooms that include kitchen, living room, bathroom, and bedroom, and more than 70 unique object categories. This challenge has two tracks: 1-Phase and 2-Phase task.

We test our approach on 2-Phase task of the AI2-THOR rearrangement challenge. 2-Phase task comprises of two stages: Walkthrough stage, where the agent is immersed in the goal setting and Unshuffle stage, where the agent is initialized in the same scene with shuffled object states and the agent is tasked to rearrange them to their initial states. For each episode there can be 1-5 objects whose state has been changed in the unshuffle task. State changes can occur either through changes in the object's position or through variations in its degree of openness. The degree openness quantifies, how far an openable object like cabinet is opened or closed. To complete a rearrangement task, the agent will have to replace all the objects whose pose has been altered and change the degree of openness of all the objects to what it was during the walkthrough stage. The number of steps an agent can take in the scene is limited to 700 for walkthrough stage and 1200 for unshuffle stage.

Training a Gaussian splat for each environment takes around 20-30 minutes, depending on the scene extend. As the size of the scene increase, the number of Gaussians needed to represent the scene also increases and since we are directly optimizing the Gaussians priors to create a Gaussian Splat of the scene, increase in the number of Gaussians increases the training time. we evaluate our pipeline on a subset of this dataset, by randomly sampling 26 episodes from the dataset. Since all the models that we use for detection and segmentation are not trained on any of the scenes from the RoomR dataset, all the results shown for our agent in \cref{tab:results_rearr} are the \textbf{zero-shot} performance of the agent.

\subsection{Evaluation Metrics}

\begin{table*}[t]
    \centering
    \scalebox{1}{
    \begin{tabular}{|c||c||c||c||c|}
        \hline
        \textbf{Method} & \textbf{\% Fixed Strict}$\uparrow$ & \textbf{\% Success}$\uparrow$ & \textbf{Misplaced}$\downarrow$ & \textbf{Energy Remaining}$\downarrow$ \\
        \hline
        \textbf{Ours} & \cellcolor{green!35}\textbf{36.35} & 3.85 & \cellcolor{green!35}\textbf{0.62} & \cellcolor{green!35}\textbf{0.63} \\
        \hline
        TIDEE \cite{sarch2022tideetidyingnovelrooms} + open-everything & 28.94 & \cellcolor{green!35}\textbf{11.70} & 0.73 & 0.71 \\
        \hline
        TIDEE \cite{sarch2022tideetidyingnovelrooms} & 11.60 & 2.40 & 0.94 & 0.93 \\
        \hline
        CAVR \cite{Liu_2024_CVPR} & 24.25 & 8.90 & 0.80 & 0.80 \\
        \hline
        MaSS \cite{trabucco2022simpleapproachvisualrearrangement} & 16.56 & 4.70 & 1.02 & 1.02 \\
        \hline
    \end{tabular}%
    }
    \captionof{table}{Comparison of results obtained by our approach with other State of the art methods in AI2-THOR rearrangement challenge dataset (RoomR dataset). \textit{Green color indicates the best result for a given metric}.}
    \label{tab:results_rearr}
\end{table*}

\begin{table*}[t]
    \centering
    \scalebox{1}{
    \begin{tabular}{|c||c||c||c||c|}
        \hline
        \textbf{Method} & \textbf{\% Fixed Strict}$\uparrow$ & \textbf{\% Success}$\uparrow$ & \textbf{Misplaced}$\downarrow$ & \textbf{Energy Remaining}$\downarrow$ \\
        \hline
        \CoolName - Hungarian Matching & \cellcolor{green!35}\textbf{36.35} & \cellcolor{green!35}\textbf{3.85} & \cellcolor{green!35}\textbf{0.62} & \cellcolor{green!35}\textbf{0.63} \\
        \hline
        \CoolName - Greedy Assignment & 30.25 & 3.85 & 0.69 & 0.71 \\
        \hline
    \end{tabular}%
    }
    \captionof{table}{Comparing the performance of our pipeline with Hungarian matching and Greedy Assignment strategy, for matching shuffled objects detected in the current setting with the goal setting. \textit{Green color indicates the best result for a given metric.}}
    \label{tab:ablation_rearr}
\end{table*}

The following are the metrics in the AI2-THOR rearrangement challenge \cite{weihs2021visualroomrearrangement}, that used to evaluate the performance of \CoolName .

\textbf{Success rate}, Indicates whether the agent has successfully completed a task or not. It is a binary value that equals 1 if all the objects in the unshuffle stage are in their goal state, at the end of the episode.

\textbf{\% Misplaced}, This is the ratio of number of objects misplaced at the end of an episode in unshuffle task to the number of objects misplaced at the start of that task. This metric can exceed 1 if the agent misplaces more objects than to begin with, during the unshuffle phase.

\textbf{\% Fixed}, This metric is the ratio of number of objects that were rearranged successfully to the number of objects in the shuffled configuration at the start of an episode in the unshuffle phase.

\textbf{\% Fixed Strict}, This metric is similar to \% Fixed, but  if any object that was already in the goal state was moved then this metric is set to zero.

\textbf{\% Energy Remaining}, is used to provide the agent with partial credit, for partially completing a task. There can be instances when the agent placed an object very close to the goal position, but the IOU of the current object with the corresponding object in the goal setting is not sufficient enough to consider it in the other metrics. This metric is computed as the ratio between the amount of energy at the end of the rearrangement task to that of the start. The energy function is a distance metric which increases from zero as the current object pose moves away from the goal pose.

\begin{table}
    \centering
    \begin{tabular}{|c||c||c|}
        \hline
        \textbf{Method} & \textbf{\% Fixed Strict} & \textbf{\% Fixed}\\
        \hline
        \CoolName (HM) & 36.34 & 38.91\\
        \hline
        \CoolName (GA) & 30.25 & 31.53\\
        \hline
    \end{tabular}
    \caption{Comparison between \% Fixed and \% Fixed Strict for \CoolName with Hungarian Matching (HM) and Greedy Assignment (GA).}
    \label{tab:fixed_comp}
\end{table}
\section{Results and Discussion}
\label{sec:results_disc}

\subsection{Overview of Results}
We compare the performance of our method with other state-of-the-art (SOTA) methods in \cref{tab:results_rearr}. \CoolName improves upon the current SOTA on \% Fixed Strict, \% Misplaced and \% Energy Remaining. Improvement to fixed strict suggests that our method was able to successfully rearrange more objects in the scene, compared to other SOTA methods. With a lower misplaced percentage than other methods, \CoolName was able to undergo rearrangement task with minimal disruption to objects that were not shuffled in the unshuffle phase. Lower value of \% energy remaining suggests that \CoolName rearranged more objects close to its goal location, without entirely completing the task.

We conducted an ablation analysis of our pipeline using two distinct object matching strategies to align object nodes in the shuffled configuration with those in the goal configuration: Hungarian algorithm and a greedy matching strategy, the results of which are presented in \cref{tab:ablation_rearr}. \CoolName with Hungarian matching algorithm outperforms the other method. This can be attributed to the optimal assignment guarantee of the Hungarian algorithm by finding pairs that minimizes the overall cost. However this comes with the cost of higher computational overhead, with a time complexity of $\mathcal{O}(n^3)$. On the other hand the greedy assignment strategy has a fast execution with time complexity of $\mathcal{O}(n)$, but provides a suboptimal solution to the bipartite matching problem. Since the RoomR dataset consists of single room environments with a relatively small area, the number of nodes (objects that needs to be rearranged) in the bipartite graph is sufficiently small, allowing the Hungarian algorithm to be used for matching without introducing significant latency.

\subsection{Higher Fixed Strict and Lower Success Rate}
A \% Fixed Strict of $36.35$ (\cref{tab:results_rearr}) indicates that \CoolName was able to rearrange $36.35\%$ of shuffled objects back to their goal configurations without altering the states of non-shuffled objects. While this percentage outperforms other methods in this dataset, the overall success rate remains significantly lower. In the rearrangement task, an episode is deemed successful if the agent can restore all shuffled objects to their goal states without disturbing those already correctly positioned. State of an object in RoomR dataset can be altered by moving it to a new location (changing its position $\mathbb{R}^3$ and orientation $SO(3)$) or by opening or closing an object to a certain degree. However, our pipeline, which matches object nodes across scenes in a category agnostic manner, does not distinguish between picking/placing an object and adjusting its openness. Therefore, our pipeline focuses exclusively on repositioning objects. This limitation means that for episodes requiring objects to be opened or closed, the success rate will be zero, even if all other objects are repositioned correctly.

\subsection{Minimal Misplacement of Objects}
Another significant advantage of \CoolName over prior approaches is its ability to complete the rearrangement task with minimal state changes in objects already at their goal configurations in the shuffled world state (objects that were never shuffled). This advantage is reflected in \cref{tab:results_rearr}, where \CoolName achieves the lowest values for both Misplaced and Energy Remaining. Furthermore, the close alignment between the \% Fixed Strict and \% Fixed values in \cref{tab:fixed_comp} suggests that the agent exhibited a very low rate of false-positive detection of objects to rearrange.

\subsection{Limitation and Future Works.} \CoolName leverages on DINOv2 to extract patchwise features from images, with a fixed patch size of 14x14. For small objects, like pencils, this patch size may be too coarse to capture fine details effectively. Additionally, using Gaussian splatting to model scenes can be memory intensive, which limits scalability. To address these issues, subsequent works may focus on utilizing a Gaussian Splatting model induced with semantic features to improve scene representation. In our current approach, the agent explores by sampling points in the unexplored traversable areas of the map until full coverage is achieved. However, this method is suboptimal, as it requires more steps than necessary. Future research may focus on developing a semantic based exploration strategy to minimize exploration costs and increase efficiency.

\section{Conclusion}
\label{sec:conclusion}

We present \CoolName, a novel approach to solve experience goal based rearrangement task. \CoolName uses 3D Gaussian Splat as a 3D scene representation for robotics. By using the ability of 3D Gaussian Splatting to render novel views faster, the agent can render images of the goal setting from any viewpoint with a virtual camera initialized in the splat. This allows the agent to freely explore the scene, with the guarantee that it knows how the environment was beforehand. Consistent and corresponding images allow us to use a dense feature matching method with DINOv2 patch features, to recognize the changes in the scene. We test our approach on the AI2-THOR rearrangement challenge benchmark and show that our approach provides improvement over the current SOTA methods. We hope that our work opens up innovative and exciting avenues in using 3D Gaussian Splatting as a world model \cite{melnik2018world} for Embodied AI \cite{konig2018embodied}.
{
    \small
    \bibliographystyle{ieeenat_fullname}
    \bibliography{main}

\begin{thebibliography}{52}
\providecommand{\natexlab}[1]{#1}
\providecommand{\url}[1]{\texttt{#1}}
\expandafter\ifx\csname urlstyle\endcsname\relax
  \providecommand{\doi}[1]{doi: #1}\else
  \providecommand{\doi}{doi: \begingroup \urlstyle{rm}\Url}\fi

\bibitem[Abou-Chakra et~al.(2024)Abou-Chakra, Rana, Dayoub, and Sünderhauf]{abouchakra2024physicallyembodiedgaussiansplatting}
Jad Abou-Chakra, Krishan Rana, Feras Dayoub, and Niko Sünderhauf.
\newblock Physically embodied gaussian splatting: A realtime correctable world model for robotics, 2024.

\bibitem[Batra et~al.(2020)Batra, Chang, Chernova, Davison, Deng, Koltun, Levine, Malik, Mordatch, Mottaghi, Savva, and Su]{batra2020rearrangementchallengeembodiedai}
Dhruv Batra, Angel~X. Chang, Sonia Chernova, Andrew~J. Davison, Jia Deng, Vladlen Koltun, Sergey Levine, Jitendra Malik, Igor Mordatch, Roozbeh Mottaghi, Manolis Savva, and Hao Su.
\newblock Rearrangement: A challenge for embodied ai, 2020.

\bibitem[Chu et~al.(2024)Chu, Ke, and Fragkiadaki]{chu2024dreamscene4ddynamicmultiobjectscene}
Wen-Hsuan Chu, Lei Ke, and Katerina Fragkiadaki.
\newblock Dreamscene4d: Dynamic multi-object scene generation from monocular videos, 2024.

\bibitem[Duan et~al.(2024)Duan, Wei, Dai, He, Chen, and Chen]{duan20244drotorgaussiansplattingefficient}
Yuanxing Duan, Fangyin Wei, Qiyu Dai, Yuhang He, Wenzheng Chen, and Baoquan Chen.
\newblock 4d-rotor gaussian splatting: Towards efficient novel view synthesis for dynamic scenes, 2024.

\bibitem[Fan et~al.(2024)Fan, Yang, Li, Li, and Zhang]{fan2024trim3dgaussiansplatting}
Lue Fan, Yuxue Yang, Minxing Li, Hongsheng Li, and Zhaoxiang Zhang.
\newblock Trim 3d gaussian splatting for accurate geometry representation, 2024.

\bibitem[Feng et~al.(2024)Feng, Feng, Shang, Jiang, Yu, Zong, Shao, Wu, Zhou, Jiang, and Yang]{feng2024gaussiansplashingunifiedparticles}
Yutao Feng, Xiang Feng, Yintong Shang, Ying Jiang, Chang Yu, Zeshun Zong, Tianjia Shao, Hongzhi Wu, Kun Zhou, Chenfanfu Jiang, and Yin Yang.
\newblock Gaussian splashing: Unified particles for versatile motion synthesis and rendering, 2024.

\bibitem[Gadre et~al.(2022)Gadre, Ehsani, Song, and Mottaghi]{gadre2022continuousscenerepresentationsembodied}
Samir~Yitzhak Gadre, Kiana Ehsani, Shuran Song, and Roozbeh Mottaghi.
\newblock Continuous scene representations for embodied ai, 2022.

\bibitem[Gao et~al.(2024)Gao, Xu, Cao, Mildenhall, Ma, Chen, Tang, and Neumann]{gao2024gaussianflowsplattinggaussiandynamics}
Quankai Gao, Qiangeng Xu, Zhe Cao, Ben Mildenhall, Wenchao Ma, Le Chen, Danhang Tang, and Ulrich Neumann.
\newblock Gaussianflow: Splatting gaussian dynamics for 4d content creation, 2024.

\bibitem[Gu et~al.(2023)Gu, Kuwajerwala, Morin, Jatavallabhula, Sen, Agarwal, Rivera, Paul, Ellis, Chellappa, Gan, de~Melo, Tenenbaum, Torralba, Shkurti, and Paull]{gu2023conceptgraphsopenvocabulary3dscene}
Qiao Gu, Alihusein Kuwajerwala, Sacha Morin, Krishna~Murthy Jatavallabhula, Bipasha Sen, Aditya Agarwal, Corban Rivera, William Paul, Kirsty Ellis, Rama Chellappa, Chuang Gan, Celso~Miguel de Melo, Joshua~B. Tenenbaum, Antonio Torralba, Florian Shkurti, and Liam Paull.
\newblock Conceptgraphs: Open-vocabulary 3d scene graphs for perception and planning, 2023.

\bibitem[Guédon and Lepetit(2023)]{guédon2023sugarsurfacealignedgaussiansplatting}
Antoine Guédon and Vincent Lepetit.
\newblock Sugar: Surface-aligned gaussian splatting for efficient 3d mesh reconstruction and high-quality mesh rendering, 2023.

\bibitem[Huang et~al.(2024{\natexlab{a}})Huang, Yu, Chen, Geiger, and Gao]{Huang_2024}
Binbin Huang, Zehao Yu, Anpei Chen, Andreas Geiger, and Shenghua Gao.
\newblock 2d gaussian splatting for geometrically accurate radiance fields.
\newblock In \emph{Special Interest Group on Computer Graphics and Interactive Techniques Conference Conference Papers ’24}. ACM, 2024{\natexlab{a}}.

\bibitem[Huang et~al.(2024{\natexlab{b}})Huang, Li, Cheng, and Yeung]{huang2024photoslamrealtimesimultaneouslocalization}
Huajian Huang, Longwei Li, Hui Cheng, and Sai-Kit Yeung.
\newblock Photo-slam: Real-time simultaneous localization and photorealistic mapping for monocular, stereo, and rgb-d cameras, 2024{\natexlab{b}}.

\bibitem[Keetha et~al.(2024)Keetha, Karhade, Jatavallabhula, Yang, Scherer, Ramanan, and Luiten]{keetha2024splatamsplattrack}
Nikhil Keetha, Jay Karhade, Krishna~Murthy Jatavallabhula, Gengshan Yang, Sebastian Scherer, Deva Ramanan, and Jonathon Luiten.
\newblock Splatam: Splat, track and map 3d gaussians for dense rgb-d slam, 2024.

\bibitem[Kerbl et~al.(2023)Kerbl, Kopanas, Leimk{\"u}hler, and Drettakis]{kerbl3Dgaussians}
Bernhard Kerbl, Georgios Kopanas, Thomas Leimk{\"u}hler, and George Drettakis.
\newblock 3d gaussian splatting for real-time radiance field rendering.
\newblock \emph{ACM Transactions on Graphics}, 42\penalty0 (4), 2023.

\bibitem[Kirillov et~al.(2023)Kirillov, Mintun, Ravi, Mao, Rolland, Gustafson, Xiao, Whitehead, Berg, Lo, Doll{\'a}r, and Girshick]{kirillov2023segany}
Alexander Kirillov, Eric Mintun, Nikhila Ravi, Hanzi Mao, Chloe Rolland, Laura Gustafson, Tete Xiao, Spencer Whitehead, Alexander~C. Berg, Wan-Yen Lo, Piotr Doll{\'a}r, and Ross Girshick.
\newblock Segment anything.
\newblock \emph{arXiv:2304.02643}, 2023.

\bibitem[Kolve et~al.(2022)Kolve, Mottaghi, Han, VanderBilt, Weihs, Herrasti, Deitke, Ehsani, Gordon, Zhu, Kembhavi, Gupta, and Farhadi]{kolve2022ai2thorinteractive3denvironment}
Eric Kolve, Roozbeh Mottaghi, Winson Han, Eli VanderBilt, Luca Weihs, Alvaro Herrasti, Matt Deitke, Kiana Ehsani, Daniel Gordon, Yuke Zhu, Aniruddha Kembhavi, Abhinav Gupta, and Ali Farhadi.
\newblock Ai2-thor: An interactive 3d environment for visual ai, 2022.

\bibitem[K{\"o}nig et~al.(2018)K{\"o}nig, Melnik, Goeke, Gert, K{\"o}nig, and Kietzmann]{konig2018embodied}
Peter K{\"o}nig, Andrew Melnik, Caspar Goeke, Anna~L Gert, Sabine~U K{\"o}nig, and Tim~C Kietzmann.
\newblock Embodied cognition.
\newblock In \emph{2018 6th International Conference on Brain-Computer Interface (BCI)}, pages 1--4. IEEE, 2018.

\bibitem[Kuhn(1955)]{https://doi.org/10.1002/nav.3800020109}
H.~W. Kuhn.
\newblock The hungarian method for the assignment problem.
\newblock \emph{Naval Research Logistics Quarterly}, 2\penalty0 (1-2):\penalty0 83--97, 1955.

\bibitem[Lei et~al.(2024)Lei, Wang, Zhou, and Li]{lei2024gaussnavgaussiansplattingvisual}
Xiaohan Lei, Min Wang, Wengang Zhou, and Houqiang Li.
\newblock Gaussnav: Gaussian splatting for visual navigation, 2024.

\bibitem[Li et~al.(2024{\natexlab{a}})Li, Liu, Zhou, Zhu, Cheng, Deng, and Wang]{li2024sgsslamsemanticgaussiansplatting}
Mingrui Li, Shuhong Liu, Heng Zhou, Guohao Zhu, Na Cheng, Tianchen Deng, and Hongyu Wang.
\newblock Sgs-slam: Semantic gaussian splatting for neural dense slam, 2024{\natexlab{a}}.

\bibitem[Li et~al.(2024{\natexlab{b}})Li, Chen, Li, and Xu]{li2024spacetimegaussianfeaturesplatting}
Zhan Li, Zhang Chen, Zhong Li, and Yi Xu.
\newblock Spacetime gaussian feature splatting for real-time dynamic view synthesis, 2024{\natexlab{b}}.

\bibitem[Liu et~al.(2024)Liu, Song, Li, Wang, and Jiang]{Liu_2024_CVPR}
Yuyi Liu, Xinhang Song, Weijie Li, Xiaohan Wang, and Shuqiang Jiang.
\newblock A category agnostic model for visual rearrangment.
\newblock In \emph{Proceedings of the IEEE/CVF Conference on Computer Vision and Pattern Recognition (CVPR)}, pages 16457--16466, 2024.

\bibitem[Lu et~al.(2024{\natexlab{a}})Lu, Zhang, Wang, Liu, Lu, and Tang]{lu2024manigaussiandynamicgaussiansplatting}
Guanxing Lu, Shiyi Zhang, Ziwei Wang, Changliu Liu, Jiwen Lu, and Yansong Tang.
\newblock Manigaussian: Dynamic gaussian splatting for multi-task robotic manipulation, 2024{\natexlab{a}}.

\bibitem[Lu et~al.(2024{\natexlab{b}})Lu, Guo, Hui, Chen, Yang, Tang, Zhu, and Dai]{lu20243dgeometryawaredeformablegaussian}
Zhicheng Lu, Xiang Guo, Le Hui, Tianrui Chen, Min Yang, Xiao Tang, Feng Zhu, and Yuchao Dai.
\newblock 3d geometry-aware deformable gaussian splatting for dynamic view synthesis, 2024{\natexlab{b}}.

\bibitem[Matsuki et~al.(2024)Matsuki, Murai, Kelly, and Davison]{matsuki2024gaussiansplattingslam}
Hidenobu Matsuki, Riku Murai, Paul H.~J. Kelly, and Andrew~J. Davison.
\newblock Gaussian splatting slam, 2024.

\bibitem[Melnik et~al.(2018)Melnik, Sch{\"u}ler, Rothkopf, and K{\"o}nig]{melnik2018world}
Andrew Melnik, Felix Sch{\"u}ler, Constantin~A Rothkopf, and Peter K{\"o}nig.
\newblock The world as an external memory: The price of saccades in a sensorimotor task.
\newblock \emph{Frontiers in behavioral neuroscience}, 12:\penalty0 253, 2018.

\bibitem[Melnik et~al.(2023)Melnik, B{\"u}ttner, Harz, Brown, Nandi, PS, Yadav, Kala, and Haschke]{melnik2023uniteam}
Andrew Melnik, Michael B{\"u}ttner, Leon Harz, Lyon Brown, Gora~Chand Nandi, Arjun PS, Gaurav~Kumar Yadav, Rahul Kala, and Robert Haschke.
\newblock Uniteam: Open vocabulary mobile manipulation challenge.
\newblock \emph{arXiv preprint arXiv:2312.08611}, 2023.

\bibitem[Meng et~al.(2024)Meng, Wu, Yin, and Zhang]{meng2024beingsbayesianembodiedimagegoal}
Wugang Meng, Tianfu Wu, Huan Yin, and Fumin Zhang.
\newblock Beings: Bayesian embodied image-goal navigation with gaussian splatting, 2024.

\bibitem[Oquab et~al.(2024)Oquab, Darcet, Moutakanni, Vo, Szafraniec, Khalidov, Fernandez, Haziza, Massa, El-Nouby, Assran, Ballas, Galuba, Howes, Huang, Li, Misra, Rabbat, Sharma, Synnaeve, Xu, Jegou, Mairal, Labatut, Joulin, and Bojanowski]{oquab2024dinov2learningrobustvisual}
Maxime Oquab, Timothée Darcet, Théo Moutakanni, Huy Vo, Marc Szafraniec, Vasil Khalidov, Pierre Fernandez, Daniel Haziza, Francisco Massa, Alaaeldin El-Nouby, Mahmoud Assran, Nicolas Ballas, Wojciech Galuba, Russell Howes, Po-Yao Huang, Shang-Wen Li, Ishan Misra, Michael Rabbat, Vasu Sharma, Gabriel Synnaeve, Hu Xu, Hervé Jegou, Julien Mairal, Patrick Labatut, Armand Joulin, and Piotr Bojanowski.
\newblock Dinov2: Learning robust visual features without supervision, 2024.

\bibitem[Qin et~al.(2024)Qin, Li, Zhou, Wang, and Pfister]{qin2024langsplat3dlanguagegaussian}
Minghan Qin, Wanhua Li, Jiawei Zhou, Haoqian Wang, and Hanspeter Pfister.
\newblock Langsplat: 3d language gaussian splatting, 2024.

\bibitem[Qiu et~al.(2024)Qiu, Yang, Zeng, and Wang]{qiu2024featuresplattinglanguagedrivenphysicsbased}
Ri-Zhao Qiu, Ge Yang, Weijia Zeng, and Xiaolong Wang.
\newblock Feature splatting: Language-driven physics-based scene synthesis and editing, 2024.

\bibitem[Radford et~al.(2021)Radford, Kim, Hallacy, Ramesh, Goh, Agarwal, Sastry, Askell, Mishkin, Clark, Krueger, and Sutskever]{radford2021learningtransferablevisualmodels}
Alec Radford, Jong~Wook Kim, Chris Hallacy, Aditya Ramesh, Gabriel Goh, Sandhini Agarwal, Girish Sastry, Amanda Askell, Pamela Mishkin, Jack Clark, Gretchen Krueger, and Ilya Sutskever.
\newblock Learning transferable visual models from natural language supervision, 2021.

\bibitem[Sarch et~al.(2022)Sarch, Fang, Harley, Schydlo, Tarr, Gupta, and Fragkiadaki]{sarch2022tideetidyingnovelrooms}
Gabriel Sarch, Zhaoyuan Fang, Adam~W. Harley, Paul Schydlo, Michael~J. Tarr, Saurabh Gupta, and Katerina Fragkiadaki.
\newblock Tidee: Tidying up novel rooms using visuo-semantic commonsense priors, 2022.

\bibitem[Sethian(1996)]{doi:10.1073/pnas.93.4.1591}
J~A Sethian.
\newblock A fast marching level set method for monotonically advancing fronts.
\newblock \emph{Proceedings of the National Academy of Sciences}, 93\penalty0 (4):\penalty0 1591--1595, 1996.

\bibitem[Sheikh et~al.(2023)Sheikh, Melnik, Nandi, and Haschke]{sheikh2023language}
Jannik Sheikh, Andrew Melnik, Gora~Chand Nandi, and Robert Haschke.
\newblock Language-conditioned semantic search-based policy for robotic manipulation tasks.
\newblock \emph{arXiv preprint arXiv:2312.05925}, 2023.

\bibitem[Shi et~al.(2023)Shi, Wang, Duan, and Guan]{shi2023languageembedded3dgaussians}
Jin-Chuan Shi, Miao Wang, Hao-Bin Duan, and Shao-Hua Guan.
\newblock Language embedded 3d gaussians for open-vocabulary scene understanding, 2023.

\bibitem[Shorinwa et~al.(2024)Shorinwa, Tucker, Smith, Swann, Chen, Firoozi, au2, and Schwager]{shorinwa2024splatmovermultistageopenvocabularyrobotic}
Ola Shorinwa, Johnathan Tucker, Aliyah Smith, Aiden Swann, Timothy Chen, Roya Firoozi, Monroe Kennedy~III au2, and Mac Schwager.
\newblock Splat-mover: Multi-stage, open-vocabulary robotic manipulation via editable gaussian splatting, 2024.

\bibitem[Trabucco et~al.(2022)Trabucco, Sigurdsson, Piramuthu, Sukhatme, and Salakhutdinov]{trabucco2022simpleapproachvisualrearrangement}
Brandon Trabucco, Gunnar Sigurdsson, Robinson Piramuthu, Gaurav~S. Sukhatme, and Ruslan Salakhutdinov.
\newblock A simple approach for visual rearrangement: 3d mapping and semantic search, 2022.

\bibitem[Waczyńska et~al.(2024)Waczyńska, Borycki, Tadeja, Tabor, and Spurek]{waczyńska2024gamesmeshbasedadaptingmodification}
Joanna Waczyńska, Piotr Borycki, Sławomir Tadeja, Jacek Tabor, and Przemysław Spurek.
\newblock Games: Mesh-based adapting and modification of gaussian splatting, 2024.

\bibitem[Weihs et~al.(2021)Weihs, Deitke, Kembhavi, and Mottaghi]{weihs2021visualroomrearrangement}
Luca Weihs, Matt Deitke, Aniruddha Kembhavi, and Roozbeh Mottaghi.
\newblock Visual room rearrangement, 2021.

\bibitem[Wolf et~al.(2024)Wolf, Bracha, and Kimmel]{wolf2024gs2meshsurfacereconstructiongaussian}
Yaniv Wolf, Amit Bracha, and Ron Kimmel.
\newblock Gs2mesh: Surface reconstruction from gaussian splatting via novel stereo views, 2024.

\bibitem[Wu et~al.(2024)Wu, Yi, Fang, Xie, Zhang, Wei, Liu, Tian, and Wang]{wu20244dgaussiansplattingrealtime}
Guanjun Wu, Taoran Yi, Jiemin Fang, Lingxi Xie, Xiaopeng Zhang, Wei Wei, Wenyu Liu, Qi Tian, and Xinggang Wang.
\newblock 4d gaussian splatting for real-time dynamic scene rendering, 2024.

\bibitem[Xiao et~al.(2024)Xiao, Wang, Li, Cai, Fan, Xue, Yang, Shen, and Gao]{xiao2024bridging3dgaussianmesh}
Yuting Xiao, Xuan Wang, Jiafei Li, Hongrui Cai, Yanbo Fan, Nan Xue, Minghui Yang, Yujun Shen, and Shenghua Gao.
\newblock Bridging 3d gaussian and mesh for freeview video rendering, 2024.

\bibitem[Xie et~al.(2024)Xie, Zong, Qiu, Li, Feng, Yang, and Jiang]{xie2024physgaussianphysicsintegrated3dgaussians}
Tianyi Xie, Zeshun Zong, Yuxing Qiu, Xuan Li, Yutao Feng, Yin Yang, and Chenfanfu Jiang.
\newblock Physgaussian: Physics-integrated 3d gaussians for generative dynamics, 2024.

\bibitem[Yan et~al.(2024)Yan, Qu, Xu, Zhao, Wang, Wang, and Li]{yan2024gsslamdensevisualslam}
Chi Yan, Delin Qu, Dan Xu, Bin Zhao, Zhigang Wang, Dong Wang, and Xuelong Li.
\newblock Gs-slam: Dense visual slam with 3d gaussian splatting, 2024.

\bibitem[Yang et~al.(2024)Yang, Yang, Pan, and Zhang]{yang2024realtimephotorealisticdynamicscene}
Zeyu Yang, Hongye Yang, Zijie Pan, and Li Zhang.
\newblock Real-time photorealistic dynamic scene representation and rendering with 4d gaussian splatting, 2024.

\bibitem[Yenamandra et~al.(2024)Yenamandra, Ramachandran, Khanna, Yadav, Vakil, Melnik, B{\"u}ttner, Harz, Brown, Nandi, et~al.]{yenamandra2024towards}
Sriram Yenamandra, Arun Ramachandran, Mukul Khanna, Karmesh Yadav, Jay Vakil, Andrew Melnik, Michael B{\"u}ttner, Leon Harz, Lyon Brown, Gora~Chand Nandi, et~al.
\newblock Towards open-world mobile manipulation in homes: Lessons from the neurips 2023 homerobot open vocabulary mobile manipulation challenge.
\newblock \emph{arXiv preprint arXiv:2407.06939}, 2024.

\bibitem[Zhong et~al.(2024)Zhong, Yu, Wu, and Li]{zhong2024reconstructionsimulationelasticobjects}
Licheng Zhong, Hong-Xing Yu, Jiajun Wu, and Yunzhu Li.
\newblock Reconstruction and simulation of elastic objects with spring-mass 3d gaussians, 2024.

\bibitem[Zhou et~al.(2024)Zhou, Chang, Jiang, Fan, Zhu, Xu, Chari, You, Wang, and Kadambi]{zhou2024feature3dgssupercharging3d}
Shijie Zhou, Haoran Chang, Sicheng Jiang, Zhiwen Fan, Zehao Zhu, Dejia Xu, Pradyumna Chari, Suya You, Zhangyang Wang, and Achuta Kadambi.
\newblock Feature 3dgs: Supercharging 3d gaussian splatting to enable distilled feature fields, 2024.

\bibitem[Zhu et~al.(2024)Zhu, Qin, Wang, Liu, and Wang]{zhu2024semgaussslamdensesemanticgaussian}
Siting Zhu, Renjie Qin, Guangming Wang, Jiuming Liu, and Hesheng Wang.
\newblock Semgauss-slam: Dense semantic gaussian splatting slam, 2024.

\bibitem[Zuo et~al.(2024)Zuo, Samangouei, Zhou, Di, and Li]{zuo2024fmgsfoundationmodelembedded}
Xingxing Zuo, Pouya Samangouei, Yunwen Zhou, Yan Di, and Mingyang Li.
\newblock Fmgs: Foundation model embedded 3d gaussian splatting for holistic 3d scene understanding, 2024.

\bibitem[Zwicker et~al.(2001)Zwicker, Pfister, van Baar, and Gross]{964490}
M. Zwicker, H. Pfister, J. van Baar, and M. Gross.
\newblock Ewa volume splatting.
\newblock In \emph{Proceedings Visualization, 2001. VIS '01.}, pages 29--538, 2001.

\end{thebibliography}
}


\end{document}